\documentclass[runningheads]{llncs}
\usepackage{graphicx}
\usepackage{hhline}
\usepackage{amsmath,amssymb,amsfonts}

\usepackage[bookmarks=false]{hyperref}
\usepackage{algorithmic}
\usepackage{textcomp}

\usepackage{framed,multirow}
\usepackage{latexsym}
\usepackage[dvipsnames]{xcolor}
\usepackage{soul}
\usepackage{xcolor}
\usepackage{hyperref}
\usepackage{color}

\usepackage{pifont}

\usepackage{enumitem}
\usepackage{threeparttable}
\usepackage{caption}
\usepackage{booktabs}

\usepackage{makecell}
\usepackage{adjustbox}
\usepackage[figuresright]{rotating}
\usepackage{tabularx}
\usepackage{cleveref}
\usepackage{bbding}
\usepackage[marginal]{footmisc}

\begin{document}
\title{DU-Net based Unsupervised Contrastive Learning for Cancer Segmentation in Histology Images}
\titlerunning{DCLR for cancer segmentation}
%
%
\author{Yilong Li \inst{2}, Yaqi Wang \inst{1}\inst{(}\Envelope\inst{)}, Huiyu Zhou \inst{3}, Huaqiong Wang  \inst{1}, Gangyong Jia \inst{4}, Qianni Zhang \inst{2}\inst{(}\Envelope\inst{)}}
\authorrunning{Yilong et al.}
%
\institute{
Communication University of Zhejiang \\
\and
Queen Mary University of London \\
\and
University of Leicester\\
\and
Hangzhou Dianzi University}

%
\maketitle              
\begin{abstract}
In this paper, we introduce an unsupervised cancer segmentation framework for histology images. The framework involves an effective contrastive learning scheme for extracting distinctive visual representations for segmentation. The encoder is a Deep U-Net (DU-Net) structure which contains an extra fully convolution layer compared to the normal U-Net.
A contrastive learning scheme is developed to solve the problem of lacking training sets with high-quality annotations on tumour boundaries. 
A specific set of data augmentation techniques are employed to improve the discriminability of the learned colour features from contrastive learning. 
Smoothing and noise elimination are conducted using convolutional Conditional Random Fields. The experiments demonstrate competitive performance in segmentation even better than some popular supervised networks.

\end{abstract}

\section{Introduction}
\label{sec:intro}
Histopathology images are considered as the gold standard for cancer diagnosis and grading. Depending on the type of cancer and the organ it is in, the assessment criteria varies, but the segmentation of tumour tissue out of the surrounding tissue is a fundamental step. This paper attempts to address the challenging problem of tumour segmenting in histopathology images. Recently, deep learning networks show superior performance in image segmentation tasks when they are appropriately trained with abundant image samples and corresponding annotation. However, if the quality and amount of annotation can not be guaranteed, the trained model's performance will be negatively impacted. This is often the case, unfortunately, due to the high requirement for time and resource in such annotation process. In addition, the inter and intra-observer variability will generally lead to unsatisfactory training outcome. Therefore, unsupervised approaches which do not rely on manual annotated training data are highly desirable. Learning effective visual representations without human supervision is a long-standing goal. Most mainstream approaches fall into one of the two classes: generative or discriminative. Generative approaches learn to generate or otherwise model pixels in the input space\cite{AFL}. In the generative approaches, pixel-level generation is computationally expensive and may not be necessary for representation learning. In comparison, discriminative methods learn representations using objective functions similar to those used for supervised learning, but train networks to perform tasks where both the inputs and labels are derived from an unlabeled dataset. Many of such approaches have relied on heuristics to design tasks\cite{clr11,clr12,clr13,clr14}, which could limit the generality of the learned representations. Discriminative approaches based on contrastive learning in the latent space have recently shown great promise, achieving state-of-the-art results\cite{clr21,clr22,clr23,clr24}. 

In this work, we introduce a histopathology image segmentation framework named Deep U-Net with Contrastive Learning of visual Representations (DCLR). The core of the framework is a contrastive learning scheme which can work in a unsupervised manner, enabling learning of effective feature representations without relying on annotated boundaries \cite{chaitanya2020contrastive,hu2021region}. This framework is capable of using a learnable nonlinear transformation between the representation and the contrastive loss can substantially improve the quality of the learned representations. Besides, the representation learning with contrastive cross entropy loss benefits from normalized embeddings. 

In this framework, the segmentation task is achieved through an unsupervised classification approach applied on patches extracted from whole slide images (WSIs). Based on the augmented data, the classification network first learns to classify the patches into tumour and non-tumour classes. The classified patches are then replaced into WSIs, and the segmentation boundaries of tumour parts are obtained. 
Finally, convolutional Conditional Random Fields (convCRFs) is applied to acquire smoother and more natural tumour region boundaries.

\section{Methods}
\subsection{Data augmentation}
Data augmentation has not been considered as a systematic way to define the contrastive prediction task. Many existing approaches define contrastive prediction tasks by changing the architecture. We show that this complexity can be avoided by performing simple random cropping and with resizing of target images\cite{clr71}. This simple design conveniently decouples the predictive task from other components such as the neural network. Broader contrastive prediction tasks can be defined by extending the family of augmentations and composing them stochastically.

Several common augmentation techniques are considered here, involving spatial/geometric transformation of data, such as cropping and resizing with flipping, rotation, and cutout. Since the operation of cropping and rotation are not significantly useful in histopathology images, this paper only considers employing resizing and cutout \cite{clr5}. The other type of augmentation involves appearance transformation, such as colour distortion, including colour dropping, modifying brightness, contrast, saturation, hue, Gaussian blurring, and Sobel filtering \cite{clr61,clr62}.
Random colour distortion is an important augmentation technique specifically to the histology images. One common problem in histology image segmentation is that most patches from the tumour regions and non-tumour regions share a similar colour distribution. Neural networks may fail to extract distinctive features from the two types of regions due to the similar colour distributions. Therefore, it is critical to use colour distortion to relieve this problem and enable learning discriminative colour features to represent different types of regions.
Strong colour augmentation substantially improves the performance of the learned unsupervised models. In this context, we argue that the simple cropping operation plus strong colour distortion can outperform some complicated methods, while it may not work as effectively in training supervised models, or even lead to negative effects. This argument is supported by experimental results presented in Section 3. 

\subsection{Network structure}

Generative approaches focus on pixel generation or modelling in the input space. However, pixel based generation is computationally expensive. 
In order to learn an effective representation with less computation cost, discriminative approaches use objective function and train networks to obtain the visual representation. Contrastive learning benefits from larger batch sizes and longer training. Discriminative approaches based on contrastive learning in the latent space show great promise, achieving state-of-the-art results in classification \cite{wang2021contrastive}. 

The proposed DCLR framework exploits contrastive learning of visual representations for histology image segmentation.
It learns discriminative representations by maximizing the agreement between differently augmented views of the same data samples. At the same time, it learns representations by minimizing the agreement between augmented views of the different data samples from different classes based on a contrastive loss in the latent space. The learning process is illustrated in Fig. 1. 
\begin{figure}[h]
	\includegraphics[width=\linewidth]{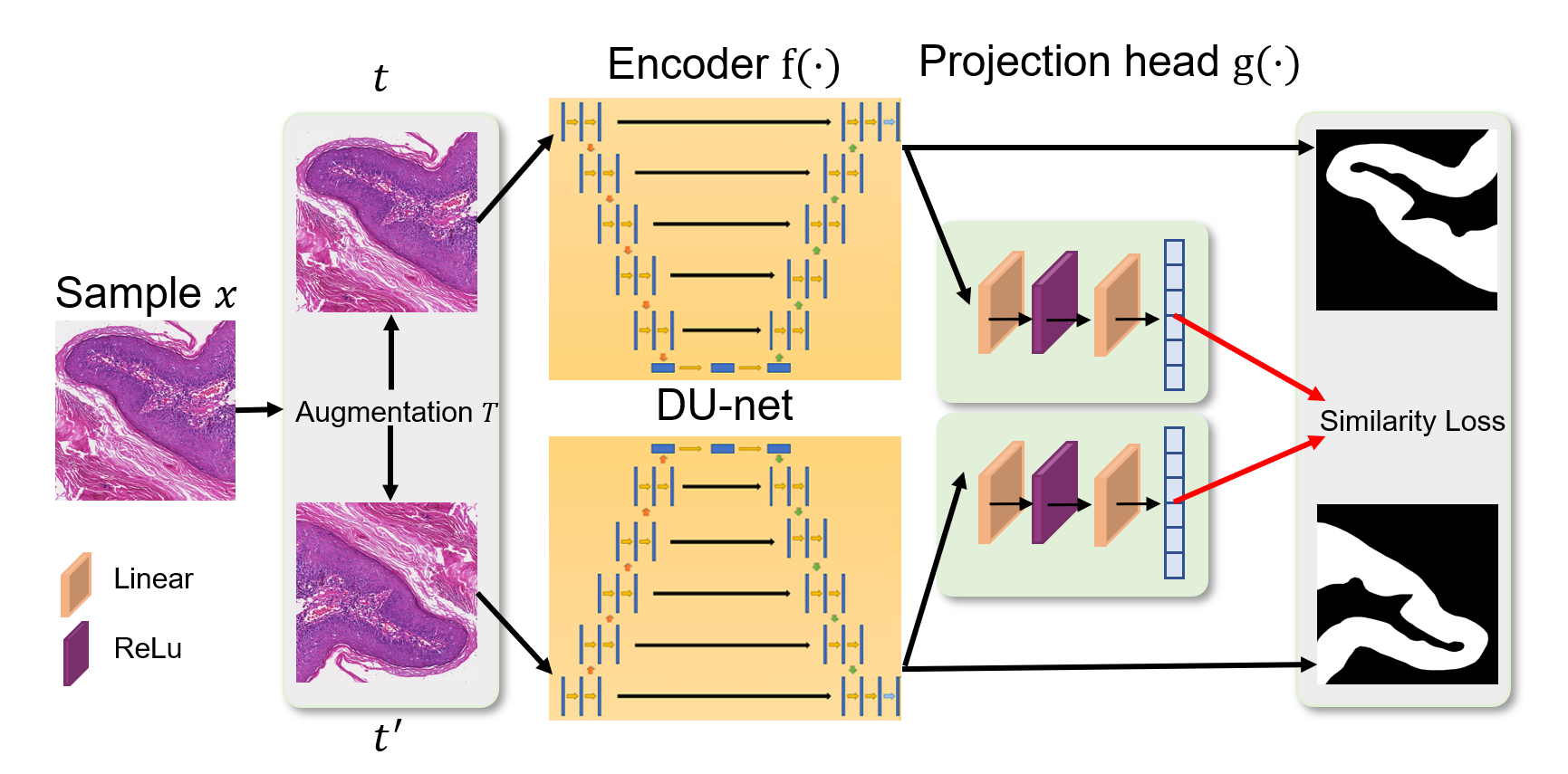}
	\caption{An illustration of the DCLR framework. Two separate data augmentation operations $t$ and $t’$, selected from a series of augmentation operations $T$, are applied to the original data sample $x$. A base encoder network $f(·)$ and a projection head $g(·)$ are trained to maximize agreement using a contrastive loss. After the training is complete, we use the encoder $f(·)$ and representation $g(·)$ for downstream tasks. }
\end{figure}

We define the batch size of the network input as $b$, and with contrastive prediction on pairs of augmented samples, the output of two augmentation operators are sized $2b$.
A stochastic data augmentation module transforms any given data sample randomly, resulting in two correlated views of the same sample, denoted $x_{i}$ and $x_{j}$, which are considered as a positive pair. 
We keep the positive pair of samples, and treat the remaining $2(b-1)$ example as negative samples. The proposed loss can guide the model to present agreement between differently augmented views of the same data sample.
DCLR framework allows various choices of the network architecture. Here, we choose $f(\cdot)=$DU-Net$(\cdot)$ to encode the $2b$ correlated views. Thus $h_{i} = f(x_{i})=$DU-Net$(x_{i})$, where $h_{i}\in R^{d}$ is the output after the average pooling layer and $d$ is the dimensions of network layers. Then a small neural network projection head $g(\cdot)$ is used in the space where contrastive loss is applied. After that, we use a Multi-Layer Perception(MLP) with one hidden layer to obtain $z_{i} = g(h_{i}) = W_{2}\sigma(W_{1}\cdot h_{i})$ where $\sigma$ is a ReLU nonlinearity, and $W_{1}$, $W_{2}$ are weights. 

In DU-Net, the separation border is computed using morphological operations. The weight map is then computed as
\begin{equation}
w(x) = w_{c}(x)+w_{0}\cdot exp(-\frac{(d_{1}(x)+d_{2}(x))^{2}}{2\sigma ^{2}})
\end{equation}

where $w_{c}:\Omega \rightarrow \mathbb{R}$ is the weight map to balance the class frequencies, $d_{1}:\Omega \rightarrow \mathbb{R}$
denotes the distance to the border of the nearest cell and $d_{2}:\Omega \rightarrow \mathbb{R}$ the distance
to the border of the second nearest cell. In our experiments we set $w_{0} = 10$ and $\sigma \approx 5$ pixels.

During training stage, the base encoder network $f(\cdot)$ and the projection head $g(\cdot)$ are constantly optimized in iterations to maximize the agreement using a contrastive loss. 
Let $s_{i,j} = z_{i}^{T}\cdot z_{j}/(\left \| z_{i} \right \|\cdot \left \| z_{j} \right \|)$
denote the dot product between normalized $z_{i}$ and $z_{j}$ (i.e. cosine similarity). Then the loss function of this network for a positive pair of examples $(i, j)$ is defined as:
\begin{equation}
L_{i,j} = -log\frac{exp(sim(z_{i},z_{j})/\tau )}{\sum_{k=1}^{2N}H_{[k\neq i]}exp(sim(z_{i},z_{j})/\tau )},
\end{equation}
where $H_{[k\neq i]}\in \{0,1\}$
is an indicator function evaluating to 1 if $k\neq i$ and $\tau$ denotes a temperature parameter. The final loss is computed across all positive pairs, both $(i, j)$ and $(j, i)$ are in a same batch. For each epoch, we update networks $f$ and $g$ to minimize $L$. After the training is complete, we discard the projection head $g(\cdot)$ and use the encoder $f(\cdot)$ and the representation $g(\cdot)$ for downstream tasks.

\subsection{Post-processing}
After the initial segmentation, a set of clear segmentation boundaries are obtained, but some noise remains. This is because the mask boundaries are mainly constructed using small patches of $512\times512$ pixels, leaving patch corners and edges in predicted boundaries. 
Moreover, the isolated tumour cells and fine details in boundaries are often not considered in human manual labeling, leading to very different ground truth boundaries compared to the predicted ones.
To approximate the manner of manual labelled boundaries, a post-processing step based on convolutional Conditional
Random Fields (CRF) is designed to adjust the obtained segmentation boundaries so that they are more inline with the hand-drawn boundaries.

All parameters of the convolutional CRFs are optimised using back propagation. 
The convolutional CRFs supplement full CRFs with a conditional independence assumption \cite{Krhen}, which assumes that the label distribution of two pixels $i$ and $j$ are conditionally independent, if the
Manhattan distance $d(i, j) > k$. Here, the hyper-parameter $k$ is defined as the filter-size.
This is a very strong assumption implying that the pairwise potential is zero, for all pairs of pixels whose distance exceed $k$. 
Consider an input $\mathbf{F}$, i.e. the probability map resulting from the unsupervised network with shape $[b, c, h, w]$, where $b$, $c$, $h$, $w$ denote batch size, number of classes, input height and width, respectively,
for a Gaussian kernel $g$ defined by feature vectors $f_{1},...,f_{d}$, for each
shape $[b, h, w]$, we define its kernel matrix as:
\begin{equation}
k_{g}[b, dx, dy, x, y]=exp(-\sum_{i=1}^{d}\frac{\Omega}{2\theta _{i}^{2}}).
\end{equation}
where $\Omega$ is defined as:
\begin{equation}
\Omega = \left | f_{i}^{(d)}[b,x,y]-f_{i}^{(d)}[b,x-dx,y-dy] \right |^{2},
\end{equation}
where $\theta_{i}$
is a learnable parameter. For a set of Gaussian kernels $\{g_1...g_s\}$ we define the merged kernel
matrix K as K$:=\sum_{i=1}^{s}w_{i}\cdot g_{i}$, $w_{i}$ is the weight of $g_{i}$. The resulting $\mathbf{Q}$ of the combined message passing of all $s$ kernels is
now given as:
\begin{equation}
\mathbf{Q}[b,c,x,y]=\sum_{dx,dy\leq k}\mathbf{K}[b,dx,dy,x,y]\cdot \mathbf{F}[b,c,x+dx,y+dy].
\end{equation}
The final output of the convolutional CRFs is the matrix $\mathbf{Q}$.

\section{Experiments}
\subsection{Dataset}

The proposed methodology is  tested on a dataset of skin cancer histopathology scans. It is relatively difficult to decide the tumour region boundaries in such images, and some types of the skin cancer present nearly no colour difference between tumour and non-tumour regions. There are 3 types of skin cancer in this dataset, basal cell cancer(BCC), squamous cell cancer(SCC) and seborrheic keratosis cancer(SKC), each consisting of about 80 WSIs. Based on a random selection, 80\% of the images are used for training and the remaining 20\% are testing. 

\subsection{Preprocessing}

To deal with WSIs in large resolution and limited annotations, we adopt the patch-based approach, which achieves WSI segmentation by firstly classifying small patches. Before patch extraction, the white background in WSIs is removed using a thresholding method on pixel intensities. 
For each patch in an annotated area, if 75\% of it is annotated as the same tissue, this patch is defined as a sample of the corresponding category. 
As discussed before, a series of augmentation operations are applied, including resizing, cutout, colour distortion, Gaussian blurring, etc. The effects of these augmentation operations are illustrated in Fig.2.
\begin{figure}[h]
	\includegraphics[width=\linewidth]{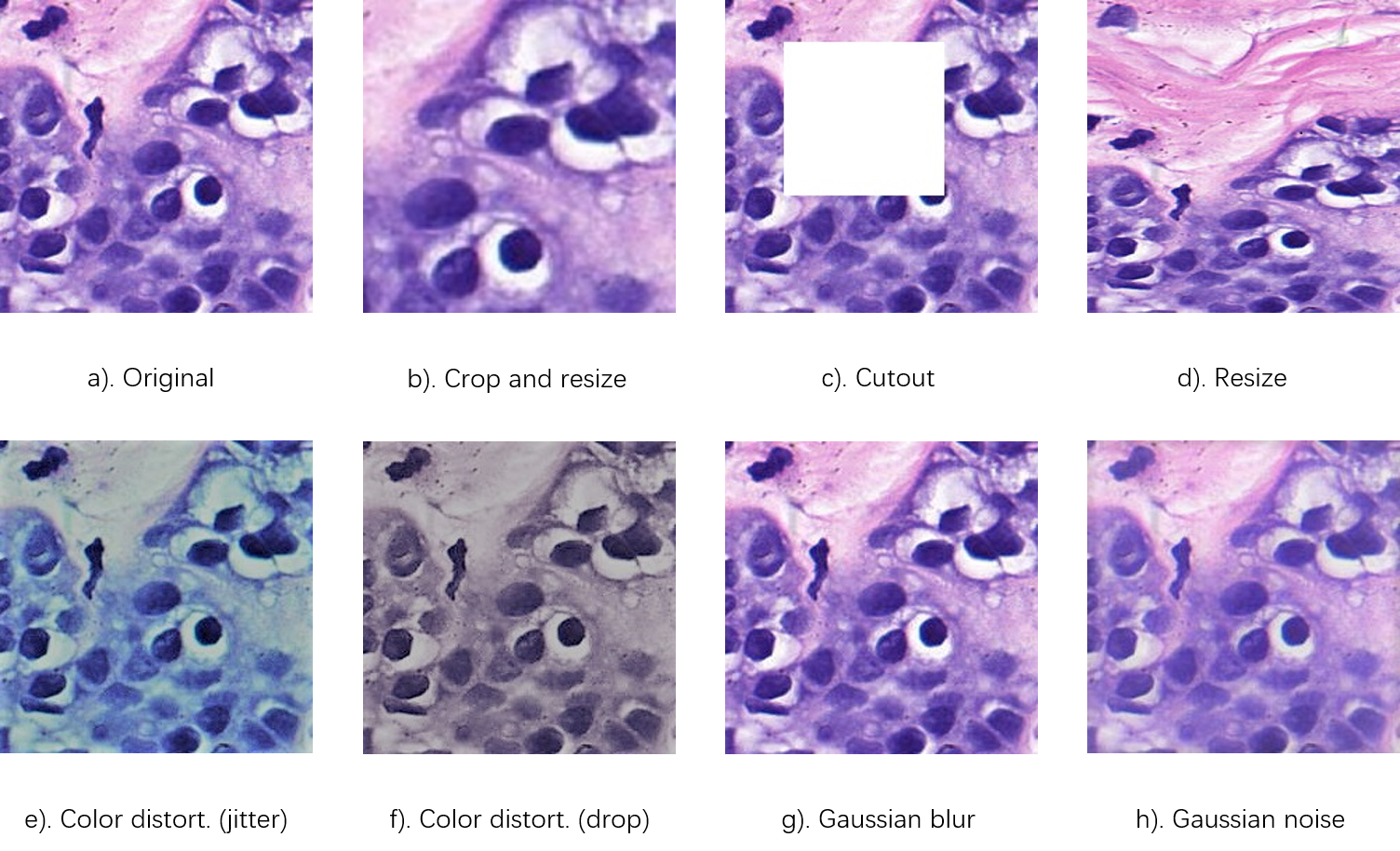}
	\caption{Data augmentations: each augmentation can transform data stochastically with some internal parameters. The augmentation policy used to train the
models includes random cropping with flipping and resizing, colour distortion, and Gaussian blurring.}
\end{figure}


\subsection{Network training details}
The model is trained using stochastic gradient descent with an adaptive learning rate. The learning rate is 0.001, and the batch size is 64, it takes around 25 minutes for one epoch. The training is terminated if the validation accuracy does not increase for 20 epochs.

\subsection{Post-processing}
After network training, the patches are placed into the whole slide image with their class labels. The resulting segmentation boundary includes obvious patch edges which influence both the segmentation accuracy and visual impressions. So the convolutional CRFs is used to smooth the boundary and eliminate the noise. Some examples of segmentation results before and after the convolutional CRFs are shown in Fig. 3.
\begin{figure}[]
\centering
	\includegraphics[width=12.6cm]{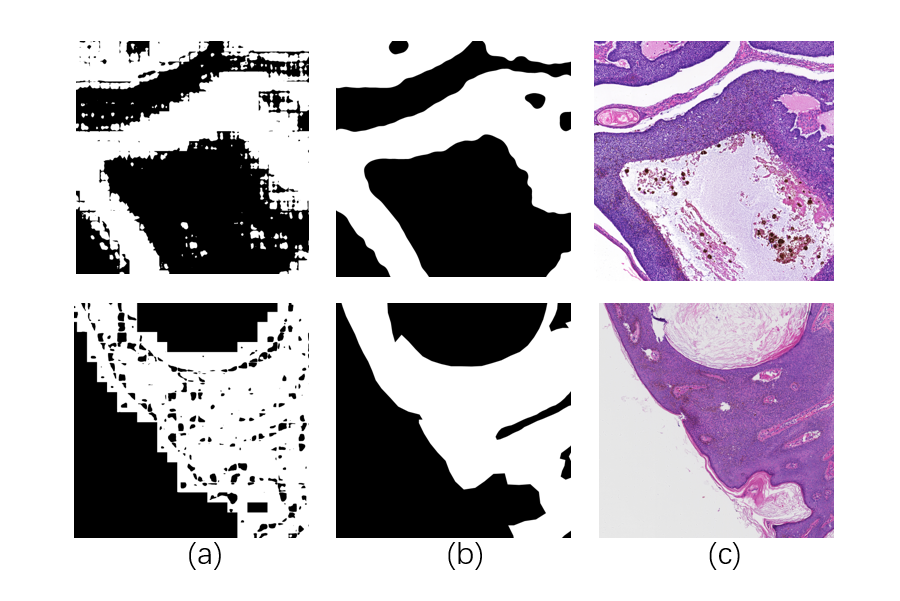}
	\caption{Segmentation result: (a) the original patch-based segmentation masks; (b) the segmentation masks after post-processing; (c) the original image}
\end{figure}

\subsection{Result evaluation}
Table 1 shows the classification results by the proposed DCLR framework compared with three popular networks, simCLR\cite{simCLR}, MoCo\cite{moco} and MoCo-v2\cite{mocov2}, on the dataset. Cases from the three types of skin cancer are treated separately. A model is trained for each cancer type.
\begin{table}[]
\centering
\setlength{\tabcolsep}{6mm}{
\begin{tabular}{ccccc}
\Xhline{3\arrayrulewidth}
\hline
\multicolumn{5}{c}{Accuracy of classification} \\\hline

dataset & simCLR & MoCo  & MoCo-v2 & DCLR(our)  \\\hline
BCC     & 93.21  & \textbf{94.94} & 94.62   & 94.81 \\
SCC     & 95.85  & 95.71 & 96.27   & \textbf{96.62} \\
SKC     & 92.14  & \textbf{94.81} & 93.90   & 93.77 \\
\hline
\Xhline{3\arrayrulewidth}
\end{tabular}}
\caption{Classification results for the skin cancer dataset and the three sets of skin cancer: basal cell cancer(BCC), Squamous cell cancer(SCC), Seborrheic keratosis cancer(SKC)}
\end{table}
It can be seen that the accuracy of the proposed unsupervised classification method is on a competitive level with that of the other popular unsupervised classification networks. The training of the DCLR framework does not require any ground truth annotation, meaning that it can work in an unsupervised manner on un-annotated datasets.

After the classification on patches, segmentation masks of WSIs are produced by placing the patches back in whole images. The segmentation results of DCLR are compared with that of the patch based approaches relying on the three classification networks, as shown in Table 2.
\begin{table}[b]
\centering
\setlength{\tabcolsep}{6mm}{
\begin{tabular}{ccccc}
\multicolumn{5}{c}{Dice Coefficient}                                     \\
\Xhline{3\arrayrulewidth}
\hline
dataset & simCLR & MoCo  & MoCo-v2 & DCLR  \\\hline
BCC           & 60.70 & 61.02  & 61.76      & \textbf{62.58} \\
SCC        & 60.53 & 58.86  & 60.88      &  \textbf{61.32} \\
SKC & 60.31 & 60.97  & \textbf{61.46}      & 61.19 \\
\hline
\Xhline{3\arrayrulewidth}
\end{tabular}}
\caption{Segmentation results for skin cancer types}
\end{table}

The segmentation based on DCLR achieves the highest Dice coefficient in two classes, BCC and SCC, while it is lower than that of MoCo-v2 in SKC dataset with a small margin, mainly due to the patched edges and corners remaining in the segmentation boundary. To enhance the segmentation boundaries and remove the unwanted patch effects, the convolutional CRF based post-processing is applied on the boundaries. Table 3 compares the output of each method after post-processing.
\begin{table*}[]
\centering
\setlength{\tabcolsep}{6mm}{
\begin{tabular}{ccccc}
\multicolumn{5}{c}{Dice Coefficient(after post-processing)}              \\
\Xhline{3\arrayrulewidth}
\hline
dataset & simCLR & MoCo  & MoCo-v2 & DCLR  \\\hline
BCC           & 68.49  & 69.34     & 69.51 & \textbf{69.76} \\
SCC        & 63.88 & 66.95  & 67.34 & \textbf{67.61} \\
SKC & 68.70 & 67.73  & \textbf{69.26} & 69.06 \\
\hline
\Xhline{3\arrayrulewidth}
\end{tabular}}
\caption{Segmentation result skin cancer after post-processing}
\end{table*}
All the results of patch based methods are improved after the post-processing. DCLR performs the best in BCC and SCC classes, while MoCo-v2 still outperforms in the SKC class, but the margin is reduced. 

\section{Conclusions}
In this paper, we present a DCLR framework that exploits the contrastive visual representation learning scheme based on a DU-Net structure for unsupervised segmentation of histopathology images. We propose a dedicated series of data augmentation operations, a modified loss function to improve the segmentation performance of this unsupervised network and a deeper U-Net to improve the accuracy of the encoder. We carefully study its components, and show the effects of different design choices on the segmentation output. By combining our proposed methods, we achieve the state of the art segmentation performance in comparison against the popular unsupervised patch based or end-to-end segmentation methods. Overall, the unsupervised learning relieves the dependence on manually annotated segmentation as an essential requirement in conventional supervised learning, and is a highly desirable feature in computational histopathology.

In the future, we aim to follow the unsupervised learning stream and further improve the segmentation quality, especially focusing on data which presents less distinctive colours in tumour and non-tumour regions. One idea is to extract features from the last convolutional layer in order to discover more effective and discriminative features in such scenarios.

\section*{Acknowledgement}
The work was partially supported by the National Key Research and Development Program under Grant No. 2019YFC0118404, the Basic Public Welfare Research Project of Zhejiang Province (LGF20H180001).

\bibliographystyle{splncs04}
\bibliography{reference}

\end{document}